\documentclass[letterpaper, 10pt, conference]{ieeeconf}      

\makeatletter
\let\NAT@parse\undefined
\makeatother

\IEEEoverridecommandlockouts                              

\let\labelindent\relax

\usepackage{graphics} 
\usepackage{subfiles}
\usepackage{csquotes}
\usepackage{gensymb}
\usepackage{pifont}
\usepackage{textcomp}
\usepackage[table]{xcolor}
\usepackage{todonotes}
\usepackage{pdfpages}
\usepackage{blindtext}
\usepackage{soul}
\usepackage{graphicx}
\usepackage{mathtools}
\usepackage{multirow}
\usepackage{booktabs}
\usepackage{flushend}
\usepackage{mwe}

\graphicspath{{images/}{../images/}}
\usepackage{subcaption}
\usepackage[font={footnotesize}, skip=5pt]{caption}
\usepackage[export]{adjustbox}
\usepackage[resetlabels]{multibib}
\usepackage{cuted}

\renewcommand{\baselinestretch}{0.975}

\usepackage{amsmath} 
\usepackage{amssymb} 
\usepackage{amsfonts}
\usepackage{bm} 
\usepackage{siunitx}
\usepackage{cancel}
\usepackage[inline]{enumitem}

\usepackage{microtype}

\newcommand{\eg}{e.\,g. }
\newcommand{\etal}{\textit{et al. }}

\newcommand{\secref}[1]{Sec.~\ref{#1}}
\renewcommand{\eqref}[1]{Eq.~(\ref{#1})}
\newcommand{\figref}[1]{Fig.~\ref{#1}}
\newcommand{\tabref}[1]{Tab.~\ref{#1}}

\newcommand{\Loss}{\mathcal{L}}

\newcommand{\veryshortarrow}[1][3pt]{\mathrel{%
   \hbox{\rule[\dimexpr\fontdimen22\textfont2-.2pt\relax]{#1}{.4pt}}%
   \mkern-4mu\hbox{\usefont{U}{lasy}{m}{n}\symbol{41}}}}
   
\newcommand{\rgbd}{\mathbf{x}}
\newcommand{\rgb}{\mathbf{I}}
\newcommand{\dpt}{\mathbf{D}}
\newcommand{\ptc}{\mathbf{P}}
\newcommand{\act}{\mathbf{a}}
\newcommand{\latent}{\mathbf{z}}
\newcommand{\context}{c}

\newcommand{\mask}{\mathbf{M}}
\newcommand{\rot}{\mathbf{R}}
\newcommand{\trn}{\mathbf{T}}
\newcommand{\sflow}{\mathbf{V}}
\newcommand{\oflow}{\mathbf{U}}
\newcommand{\proj}{\mathbf{w}}
\newcommand{\occ}{\hat{\mathbf{O}}}

\newcommand{\dist}{\mathbf{J}}
\newcommand{\vae}{q}
\newcommand{\gen}{G}

\newcommand{\rgbdx}[2]{\rgbd_{#1:#2}}
\newcommand{\actx}[2]{\act_{#1:#2}}
\newcommand{\latentx}[2]{\latent_{#1:#2}}

\newcommand{\ourmodel}{T3VIP }

\usepackage[hidelinks]{hyperref}
\usepackage{xcolor}
\hypersetup{
	colorlinks,
	linkcolor={red!50!black},
	citecolor={blue!50!black},
	urlcolor={blue!80!black}
}
\usepackage{cleveref}
\usepackage[table]{xcolor}

\title{\LARGE \bf T3VIP: Transformation-based 3D Video Prediction}

\author{Iman Nematollahi\textsuperscript{1}, Erick Rosete-Beas\textsuperscript{1}, Seyed Mahdi B. Azad\textsuperscript{1}, Raghu Rajan\textsuperscript{1}, \\ Frank Hutter\textsuperscript{1,2}, Wolfram Burgard\textsuperscript{3}
\thanks{\textsuperscript{1}University of Freiburg. \textsuperscript{2}Bosch Center for AI. \textsuperscript{3}University of Technology Nuremberg. This work has been supported by German Federal Ministry of Education and Research under contract number 01IS18040B-OML and also by the Deutsche Forschungsgemeinschaft under grant number 417962828.\looseness=-1}
}

\begin{document}

\maketitle
\thispagestyle{empty}
\pagestyle{empty}

\begin{abstract}
For autonomous skill acquisition, robots have to
learn about the physical rules governing the 3D world dynamics
from their own past experience to predict and reason
about plausible future outcomes. To this end, we propose a transformation-based 3D video prediction (T3VIP) approach that explicitly models the 3D motion by decomposing a scene into its object parts and predicting their corresponding rigid transformations. Our model is fully unsupervised, captures the stochastic nature of the real world, and the observational cues in image and point cloud domains constitute its learning signals. To fully leverage all the 2D and 3D observational signals, we equip our model with automatic hyperparameter optimization (HPO) to interpret the best way of learning from them. To the best of our knowledge, our model is the first generative model that provides an RGB-D video prediction of the future for a static camera. Our extensive evaluation with simulated and real-world datasets demonstrates that our formulation leads to interpretable 3D models that predict future depth videos while achieving on-par performance with 2D models on RGB video prediction. Moreover, we demonstrate that our model outperforms 2D baselines on visuomotor control. Videos, code, dataset, and pre-trained models are available at \url{http://t3vip.cs.uni-freiburg.de}.\looseness=-1
\end{abstract}

\section{Introduction}\label{sec:intro}
\begin{figure}[t]
	\centering
	\vspace*{-4mm}
	\includegraphics[width=1\columnwidth]{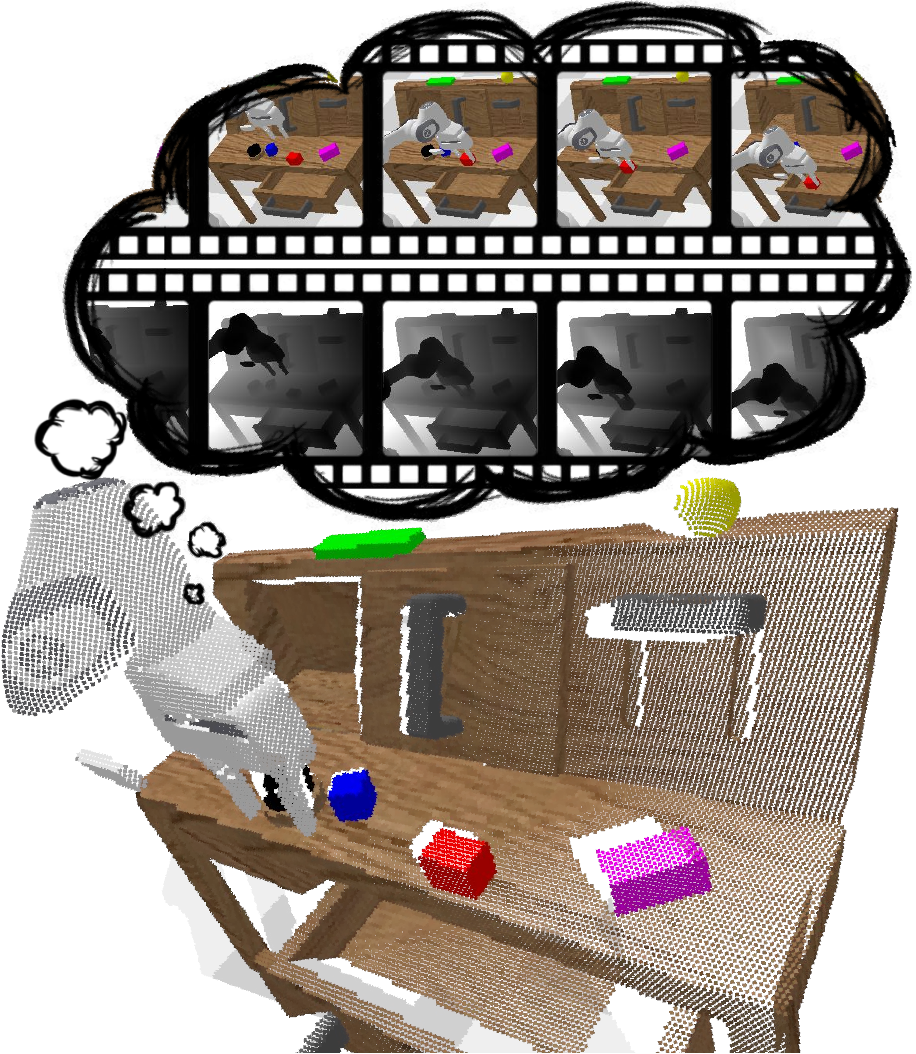}
	\caption{T3VIP learns a 3D world model from past experience to imagine plausible future RGB-D videos and plan the best action trajectory.}
	\label{fig:cover}
	\vspace*{-8mm}
\end{figure}

From an early age, humans develop an intuitive understanding of physics, capable of predicting the dynamics of the 3D world~\cite{ullman2018learning}. This cognitive model, which is learned via our interactions in the real world, enables us to generalize from past experience and predict the future observations of a novel scene through reasoning about the motion in 3D space. Furthermore, by equipping us to foresee the consequences of our actions, this mental model allows us to decide how to interact with the world and manipulate unseen objects towards our desired goals. Similarly, for a robot collecting unlabeled RGB-D video sequences autonomously, the ability to comprehend the dynamics of the surrounding 3D world and predict its likely future developments is of high value for acquiring real-world manipulation skills (see~\figref{fig:cover}). This, however, stands in contrast to the existing state-of-the-art world models, which, despite recent progress~\cite{finn2016unsupervised,ha2018world,hafner2019dream}, lack an understanding of the underlying 3D world and consider the agent's observations of the world as 2D images. Although recent 2D video prediction models~\cite{finn2016unsupervised,babaeizadeh2018stochastic,lee2018stochastic} learn to reason about motion from raw images and predict large-range pixel motion, they lack geometric scene understanding and cannot explicitly predict the 3D rigid body transformation of objects in the scene. Moreover, a robot equipped with a 2D world model observes the world in flat images and cannot disambiguate between world states with similar 2D visual appearance but different geometric structures. For example, such a robot cannot distinguish the physical dynamics of a flat disk from a sphere with similar textures.\looseness=-1

Our goal in this paper is to develop a 3D world model for autonomous agents based on their past unlabeled experience. We consider the problem of 3D video prediction, where, given a sequence of previous RGB-D frames, a generative model aims to capture spatial and temporal representations of the world and predict the RGB-D frames of the future. To this end, we propose \textbf{T}ransformation-based \textbf{3}D \textbf{Vi}deo \textbf{P}rediction (\textbf{T3VIP}), a model that decomposes a scene into salient objects and predicts their corresponding 3D rigid body transformations. Our model is fully unsupervised and learns the physical dynamics of the scene by reasoning about the visual and geometric cues elicited from the environment. We equip our model with an automatic hyperparameter optimization technique to find the best strategy for exploiting available observational signals to achieve higher prediction quality and, at the same time, reduce the dependency on human expertise. Our structured formulation allows T3VIP to compute the 3D scene flow, the 2D optical flow, and the occlusion mask as emergent properties, leading to better interpretability of the dynamics models. Thus, T3VIP is a 3D-aware world model that captures the geometric features of the world and predicts multiple future RGB-D frames. We evaluate T3VIP on one real-world and two simulated RGB-D video datasets and compare it against baselines that predict future frames solely in the pixel space. Our extensive qualitative and quantitative evaluations demonstrate that our model effectively predicts future RGB-D videos, outperforming 2D baselines on 3D servoing via model-predictive control.

The main contributions of this paper are:
1) a stochastic 3D video prediction model making long-range predictions via reasoning about rigid transformations of the scene in both action-conditioned and action-free settings, 2) an unsupervised learning framework for predicting the dynamics of a scene solely based on unlabeled 3D point clouds and 2D images, 3) the utilization of automatic hyperparameter optimization to fully leverage all available observational cues and reduce the need for human expertise, and 4) an interpretable 3D world model outperforming 2D models on model-predictive control.

\section{Related Work}\label{sec:related}
\textbf{3D Dynamics:} Learning an intuitive physics model that reasons about the motion of objects is tightly coupled with physical interactions and geometric scene understanding and, therefore, has been a long-standing goal in both robotics and vision communities. Many prior works infer the dynamics of objects solely based on RGB images or videos~\cite{janner2018reasoning,ye2018interpretable}. Other approaches leverage Newtonian physics and neural networks to predict 3D motion trajectories of objects given a single static RGB image~\cite{mottaghi2016newtonian,DBLP:journals/corr/MottaghiRGF16}. Compared to these approaches, we directly take the agent's RGB-D observations of the environment into account to reason about geometry. Recently, some approaches have considered a more general 3D setting, where an agent obtains depth information of the scene to model the forward dynamics~\cite{byravan2017se3,byravan2018se3,xu2020learning}. These methods learn the 3D rigid body motion of objects from raw point cloud data. Although they show impressive results in modeling the 3D dynamics, they require ground-truth point-wise data associations as supervision. Our prior work~\cite{nematollahi2020hindsight} proposed an approach to learn jointly forward and inverse 3D dynamics from unlabeled point clouds and images. Hind4sight-Net~\cite{nematollahi2020hindsight} relaxes the requirement of ground-truth point-wise data associations by reasoning about observational changes in 2D and 3D domains. Unlike Hind4sight-Net, which predicts a point cloud transformation for only one time step, we now model long-term 3D dynamics of the scene by predicting its future RGB-D videos.\looseness=-1

\textbf{Video Prediction:} Recent progress in generative models has led to impressive developments in video prediction models~\cite{xie2018few,tian2020model}. More related to our work, Finn \etal\cite{finn2016unsupervised} proposed a deterministic generative model which predicts pixel transformations from the previous image to form the next image. Several works have been built on this technique to address stochasticity of the real world~\cite{babaeizadeh2018stochastic}, blurry predictions~\cite{lee2018stochastic}, occlusions~\cite{ebert2017self}, and fast adaptation to unseen objects~\cite{yen2020experience}. Our work is most similar to SV2P~\cite{babaeizadeh2018stochastic}, which utilizes the CDNA~\cite{finn2016unsupervised} architecture to perform deterministic next frame generation based on random stochastic latent codes sampled from a prior distribution. The main difference between these 2D models and our approach is that we consider the RGB-D observations and explicitly reason about the 3D transformations of objects to predict future RGB-D video frames.\looseness=-1

\textbf{Model-based Reinforcement Learning:} Learning a world model~\cite{ha2018world,hafner2019dream,hafner2020mastering} from the experience of an agent to predict plausible futures and plan actions accordingly falls under the category of model-based reinforcement learning (RL). Model-based RL algorithms are generally known to surpass the data efficiency of model-free methods~\cite{deisenroth2013survey}. Several works have recently employed visual model-based RL to solve real-world robotic manipulation tasks~\cite{zhang2019solar,kalashnikov2021mt}. However, these algorithms rely solely on RGB observations of the world and do not reason about the 3D scene dynamics. In this work, we adopt a similar strategy as Finn and Levine~\cite{finn2017deep} and use our learned generative model for visual model-predictive control to showcase its capability in comprehending 3D dynamics and leveraging it to reach user-defined target points.\looseness=-1
\begin{figure*}[t!]
    \centering
    \includegraphics[scale=0.785]{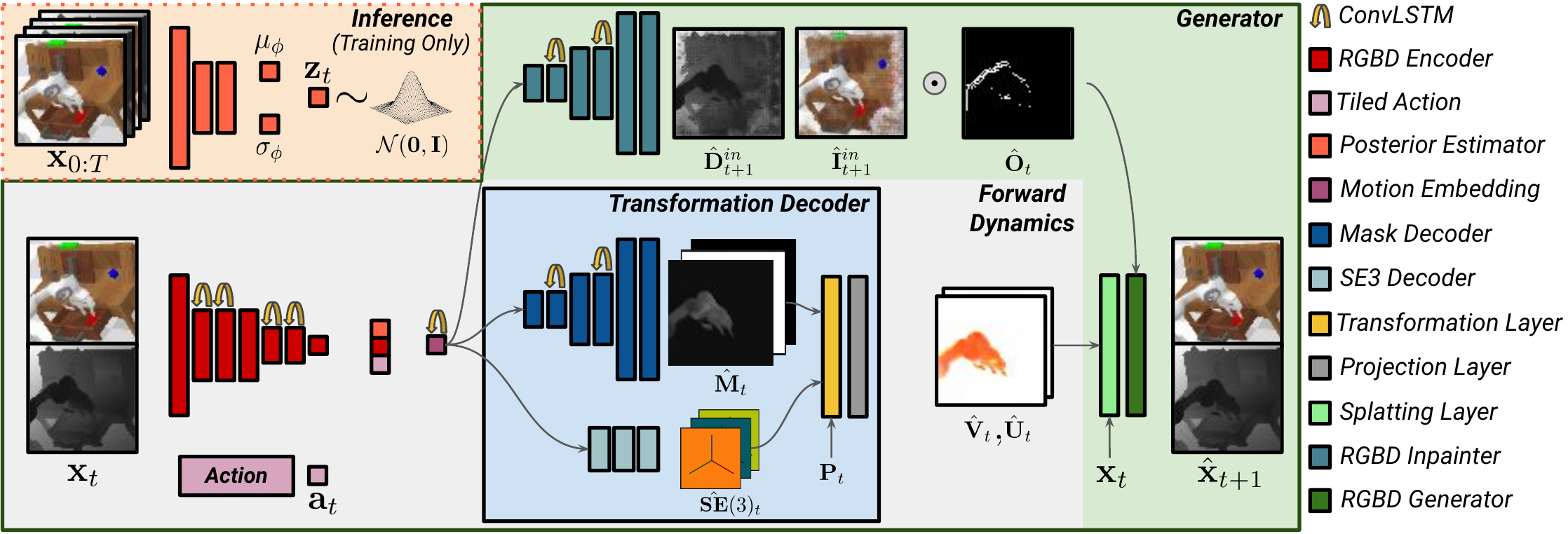}
    \caption{Structure of T3VIP: Given an RGB-D frame $\rgbd_t$, a latent variable $\latent_t$, and optionally an action $\act_t$, our generator network segments a scene into $K$ object masks and predicts their 3D transformations. The \textit{Transformation Layer} moves the initial point cloud $\ptc_t$ according to the predicted 3D motion and analytically computes scene flow $\hat{\sflow}_t$, optical flow $\hat{\oflow}_t$, and a binary occlusion mask $\occ_t$. The \textit{Splatting Layer} forward-warps the initial frame according to the predicted flow. Finally, the \textit{RGB-D Generator} takes the forward-warped frame, fills the occluded regions with the aid of the \textit{RGB-D Inpainter}, and generates the next RGB-D frame $\hat{\rgbd}_{t+1}$. During training, the latent variable $\latent_t$ is sampled from the posterior $\mathcal{N}(\mu_\phi(\rgbdx{0}{T}), \sigma_\phi(\rgbdx{0}{T}))$, which is estimated via the inference network, while at test time, it is sampled from a unit Gaussian ${\mathcal{N}(\mathbf{0}, \mathbf{I})}$. Our model is learned end-to-end in an unsupervised fashion.\looseness=-1}
    \label{fig:architecture}
    \vspace{-0.5cm}
\end{figure*}

\section{Problem Formulation}\label{sec:problem}
We aim to learn a 3D video prediction model that enables an autonomous robot to comprehend the spatiotemporal dynamics of its environment and anticipate possible futures in 3D space. We are interested in a structured generative model with an object-centric representation that explicitly predicts the 3D transformations of rigid bodies while staying invariant to their appearance. To learn the real-world dynamics, we approach this problem in an unsupervised fashion, where the agent has only access to the observational changes in the environment to learn about the motion and consequences of its actions. Moreover, we aim for a model that also captures the stochastic nature of the real world.\looseness=-1

Inspired by prior work on 2D video prediction~\cite{finn2016unsupervised}, we define the problem of RGB-D video prediction as follows: a frame $\rgbd_t = (\rgb_t, \dpt_t) \in \mathbb{R}^{4 \times H \times W}$ consists of an RGB image $\rgb_t \in \mathbb{R}^{3 \times H \times W}$ and a depth map $\dpt_t \in \mathbb{R}^{1 \times H \times W}$, where $H$ and $W$ are the height and width of the frame. 
Given the first $\context$ frames of an RGB-D video $\rgbd_0,\dots,\rgbd_{c-1}$ as \emph{context} frames  (typically $\context=2$), our goal is to predict the future frames by sampling from $p(\rgbdx{c}{T}| \rgbdx{0}{c-1})$. Following prior work on variational methods for modeling stochasticity~\cite{babaeizadeh2018stochastic}, this deterministic generative model can also be conditioned on a sequence of stochastic latent codes $\latentx{0}{T-1}$ distributed according to a prior distribution $p(\latent)$, to account for non-observable variables (such as friction or weight of objects) that might affect the future. Furthermore, to learn a 3D world model effective for model-based visuomotor control, the future prediction can be additionally conditioned on the trajectory of actions $\actx{0}{T-1}$ that the robot plans to take.\looseness=-1 

\section{\ourmodel}\label{sec:approach}
Our model consists of two components: a recurrent generator network $\gen_{\theta}$ and a unit Gaussian prior distribution $p(\latent)$. First, we explain how T3VIP explicitly models the 3D scene dynamics to generate the next RGB-D frame. After that, we describe how our prior distribution accounts for the stochasticity at test time by employing a separate inference network $\vae_{\phi}$ used only during training. Next, we detail how we utilize observational cues to learn from unlabeled data. Finally, we elaborate on how to automate the determination of the optimal hyperparameter configurations of our model. The architecture of our model is shown in \figref{fig:architecture}.\looseness=-1
\subsection{3D Forward Dynamics}\label{sec:approach_dyna}
At the core of our model lies a transformation prediction module. Given an RGB-D frame $\rgbd_t = (\rgb_t, \dpt_t)$, a latent variable $\latent_t$,  and optionally an $n$-dimensional continuous action $\act_t$ as inputs, T3VIP learns a motion embedding from which it decomposes a scene into $K$ object masks $\mask_t$ (including background) and predicts a 3D rigid body transformation $[\rot_t,\trn_t] \in \mathbf{SE}(3)$ per object. Using camera intrinsics, our model first converts the depth map $\dpt_t$ of the current frame into an ordered 3D point cloud $\ptc_t=(X_t, Y_t, Z_t)$, where each point contains the 3D coordinates of the scene, and then moves this point cloud according to the predicted object masks and $\mathbf{SE}(3)$ transformations to generate a transformed point cloud:\looseness=-1
\begin{equation}\label{eq:transformed_pointcloud}
\hat{\ptc}_{t+1} = \sum_{k=1}^{K} \mask_{t}^{k} (\rot_t^{k} \ptc_{t} + \trn_t^{k}).
\end{equation}
Since our transformation module explicitly learns the 3D motion of point cloud segments and utilizes information from the previous point cloud to construct the transformed point cloud, it remains invariant to the visual appearance of objects. Please note that $\hat{\ptc}_{t+1}$ is no longer an ordered point cloud since the transformation layer moves each point in 3D and the previous x/y-axis ordering of points are not valid anymore. Hence, we do not directly obtain the next depth map $\hat{\dpt}_{t+1}$, but a transformed depth map $\hat{\dpt}^{\prime}_{t+1}$ that is unordered. 
Nevertheless, we can compute a binary occlusion mask $\occ_t \in \mathbb{R}^{1 \times H \times W}$ analytically by performing a perspective projection test of the transformed point cloud. This mask addresses points that are occluded in the initial frame but will be visible in the next frame. Furthermore, we can compute the scene flow $\hat{\sflow}_t \in \mathbb{R}^{3 \times H \times W}$ which is the 3D motion of scene points with respect to the camera and optical flow $\hat{\oflow}_t \in \mathbb{R}^{2 \times H \times W}$ which is the projection of 3D motion onto the image plane as follows:\looseness=-1
\begin{equation}
\renewcommand*{\arraystretch}{1.3}
\hat{\sflow}_{t} = \hat{\ptc}_{t+1} - \ptc_t
\quad\text{and}\quad
\hat{\oflow}_{t} = \proj(\hat{\sflow}_{t}),
\end{equation}
where $\proj$ is the projection layer, which utilizes the camera intrinsics to project the scene flow over the camera plane. \looseness=-1

Our forward dynamics model closely resembles the forward model of Hind4sight-Net~\cite{nematollahi2020hindsight}, with the difference that T3VIP employs convolutional LSTMs to leverage the spatial invariance of frame representations for multi-step future prediction. Compared to CDNA~\cite{finn2016unsupervised} proposed by Finn \etal for 2D video prediction, our model makes the same architectural choices for encoding the frames and actions and decoding the masks of the objects, but it differs in how it decodes the motion. T3VIP predicts 3D rigid body transformations instead of 2D normalized convolution kernels.\looseness=-1
\subsection{Generator}\label{sec:approach_gen}
To generate the next RGB-D frame, our model first uses the predicted optical flow $\hat{\oflow}_t$ to forward-warp the transformed depth map $\hat{\dpt}^{\prime}_{t+1}$ and the RGB image $\rgb_t$ to the coordinates aligned with the predicted 3D motion:\looseness=-1
\begin{equation}
\renewcommand*{\arraystretch}{1.3}
\hat{\rgb}^{fw}_{t+1} = \overrightarrow{\sigma} (\rgb_t, \hat{\oflow}_t)
\quad\text{and}\quad
\hat{\dpt}^{fw}_{t+1} = \overrightarrow{\sigma} (\hat{\dpt}^{\prime}_{t+1}, \hat{\oflow}_t),
\end{equation}
where $\overrightarrow{\sigma}$ is the splatting operator proposed by~\cite{niklaus2020softmax}, and $\hat{\rgb}^{fw}_{t+1}$ and $\hat{\dpt}^{fw}_{t+1}$ are the forward-warped RGB image and depth map. While we now have the next frame generated via the predicted 3D motion, we need to address occluded regions that are missing in the new frame. To this end, our model employs an RGB-D inpainter, which takes the motion embedding as input and predicts an RGB image $\hat{\rgb}^{in}_{t+1}$, and a depth map $\hat{\dpt}^{in}_{t+1}$. Finally, T3VIP generates the next RGB-D frame $\hat{\rgbd}_{t+1}$ as follows:\looseness=-1
\begin{align} \begin{split}
    \hat{\rgb}_{t+1} &= (1 - \occ_t) \cdot  \hat{\rgb}^{fw}_{t+1} + \occ_t \cdot  \hat{\rgb}^{in}_{t+1},
\end{split} \\ \begin{split}
    \hat{\dpt}_{t+1} &= (1 - \occ_t) \cdot  \hat{\dpt}^{fw}_{t+1} + \occ_t \cdot  \hat{\dpt}^{in}_{t+1}.
\end{split} \end{align}
Our binary occlusion mask makes sure that the model mainly uses the predicted 3D transformations to generate the next frame and only occluded points of the scene get inpainted.\looseness=-1
\subsection{Modeling Stochasticity}\label{sec:approach_sto}
To enable our model to grasp the stochastic nature of the real world during training, we introduce a latent variable $\latent_t$ into our recurrent generative model and sample from $p(\rgbdx{c}{T}\mid \rgbdx{0}{c-1}, \latentx{0}{T-1})$. This hinders the generative model from directly maximizing the likelihood of the data due to the latent's dependency on $p(\rgbdx{0}{T})$. To overcome this problem, we follow the approach proposed by Babaeizadeh \etal~\cite{babaeizadeh2018stochastic} to approximate the posterior with an inference network $\vae_{\phi}(\latent_t \mid \rgbdx{0}{T})$ and optimize the variational lower bound of the log-likelihood. In order to encourage the latent variable to discover the stochastic information between frames during training, the inference network takes the entire video sequence as input and outputs the parameters of a conditionally Gaussian distribution $\mathcal{N}(\mu_\phi(\rgbdx{0}{T}), \sigma_\phi(\rgbdx{0}{T}))$. However, at test time, we sample the latents from an assumed prior which in our case is a fixed unit Gaussian ${\mathcal{N}(\mathbf{0}, \mathbf{I})}$. The optimization of our model involves minimizing the reconstruction loss between the predicted and ground-truth frames and the KL divergence between the approximated posterior and the assumed prior.\looseness=-1
\begin{equation}
  \Loss_{\mathit{rec}}(\rgbdx{0}{T}) = \mathbb{E}_{\substack{\rgbdx{0}{T} \\ 
                  \latent_{t} \sim {q_{\phi}}}} \left[\sum_{t=0}^{T-1} ||\rgbd_t - G_{\theta}(\rgbd_0, \latent_{0:t-1})||_{\alpha} \right],
\end{equation}
\begin{equation}
 \Loss_{\mathit{kl}}(\rgbd_{0:T}) = \mathbb{E}_{\rgbdx{0}{T}}\left[\sum_{t=0}^{T-1} D_{\mathit{KL}}\left(q_{\phi}(\latent_t \mid \rgbdx{0}{T}) \mid\mid p(\latent)\right)\right],
\end{equation}
where $\alpha$ is a hyperparameter specifying if the pixel-wise absolute ($\alpha=1$) or squared ($\alpha=2$) error is used to reconstruct frames. Our inference network $\vae_{\phi}$ closely mimics SV2P~\cite{babaeizadeh2018stochastic}, with the difference that ours incorporates RGB-D sequences of the video instead of RGB.\looseness=-1
\subsection{3D Point Cloud Alignment Loss}\label{sec:approach_3d}
Our approach learns the 3D scene dynamics of the real-world without requiring any external trackers and by solely relying on unlabeled RGB-D videos. In this regard, our optimization formulation leverages observational cues to enforce an explicit geometric constraint on predicting the forward dynamics. Concretely, we impose consistency between the transformed $\hat{\ptc}_{t+1}$ and observed $\ptc_{t+1}$ point clouds at each time step. Since the transformed point cloud is unordered, we employ the k-Nearest-Neighbors (kNN) algorithm to find the point-wise data association between the transformed and observed point clouds. Hence, kNN takes as input two point clouds $\hat{\ptc}_{t+1}$ and $\ptc_{t+1}$, and for each point in each point cloud, it first finds the nearest neighbor in the other point set based on their distance to the camera and then forms a distance transform by summing up the euclidian distance of corresponding points. Since kNN is not necessarily a symmetric function, we calculate the distance transforms in both directions. We define the distance transforms between the point sets as follows:\looseness=-1
\begin{align} \begin{split}
    \dist_{t+1}^{\hat{\ptc} \veryshortarrow \ptc} &= \min_{x^{\prime},y^{\prime}} \sum_{x,y} \|\hat{\ptc}_{t+1}^{xy} - \ptc_{t+1}^{x^{\prime}y^{\prime}}\|_{2},
\end{split} \\ \begin{split}
    \dist_{t+1}^{\ptc \veryshortarrow \hat{\ptc}} &= \min_{x^{\prime},y^{\prime}} \sum_{x,y} \|\ptc_{t+1}^{xy} - \hat{\ptc}_{t+1}^{x^{\prime}y^{\prime}}\|_{2},
\end{split} \end{align}
where $x$ and $y$ are the grid coordinates of points, $x^{\prime} \in (x-k, x+k)$ and $y^{\prime} \in (y-k, y+k)$ are the coordinates of the data associated points, and $k$ is the kNN parameter. Finally, we sum the distance transforms to define the point cloud alignment loss:
\begin{equation}
\label{eq:knn_dist}
\Loss_{\mathit{knn}}(\rgbd_{0:T}) =  \mathbb{E}_{\rgbdx{0}{T}}\left[\sum_{t=0}^{T-1} \dist_{t+1}^{\hat{\ptc} \veryshortarrow \ptc} + \dist_{t+1}^{\ptc \veryshortarrow \hat{\ptc}}\right].
\end{equation}
Please note that compared with Hind4sight-Net~\cite{nematollahi2020hindsight}, which utilizes a chamfer distance for the point cloud alignment loss, our kNN formulation is computationally more efficient since it finds the associated point in a smaller $k \times k$ window instead of an $H \times W$ window required by the chamfer distance.\looseness=-1
\subsection{Edge-aware Smoothness Loss}\label{sec:approach_edge}
T3VIP explicitly reasons about the 3D transformations of the scene segments in a self-supervised fashion and outputs the scene flow and optical flow as emergent properties. As an additional proxy loss for our self-supervised framework, we adopt an edge-aware second-order smoothness regularization~\cite{tomasi1998bilateral}, to encourage piecewise smoothness of geometry and motion. Our smoothness loss term measures the difference between spatially neighboring points in the scene and optical flow, adaptively weighted by the image gradients:
\begin{equation*}
\Loss_{\mathit{fs}}(\rgbd_{0:T}) =\mathbb{E}_{\rgbdx{0}{T}} \left[\sum_{t=0}^{T-1} \sum_{x,y} {\big\lvert}{\nabla^{2} {\mathbf{F}}_{t}^{xy}}{\big\lvert} \times \exp{\left(-|\nabla \rgb_{t}^{xy}|\right)} \right],
\end{equation*}
where $|\cdot|$ denotes element-wise absolute value, $\nabla$ is the vector differential operator, and ${\mathbf{F}}_{t}$ could be either the scene or optical flow. Intuitively, by utilizing an edge-aware smoothness penalty, we leverage the observation that motion boundaries and edges present in the image usually coincide.\looseness=-1

\subsection{Full Model}\label{sec:approach_full}
Our full model combines all the above-mentioned objectives to learn a stochastic 3D video prediction model from unlabeled RGB-D data. Namely, our model aims to reconstruct RGB images ($\Loss_{rec}^{I}$) and depth maps ($\Loss_{rec}^{D}$), enforce the consistency of predicted and observed point clouds ($\Loss_{knn}$), encourage scene flow ($\Loss_{fs}^{s}$) and optical flow smoothness ($\Loss_{fs}^{o}$) and fit the prior distribution ($\Loss_{kl}$). Hence, the full objective of our T3VIP model is:\looseness=-1
\begin{equation*}
\label{eq:full_loss}
\Loss = \lambda_{1}\Loss_{\mathit{rec}}^{I} + \lambda_{2}\Loss_{\mathit{rec}}^{D} + \lambda_{3}\Loss_{\mathit{knn}} + \lambda_{4}\Loss_{\mathit{fs}}^{s} + \lambda_{5}\Loss_{\mathit{fs}}^{o} + \lambda_{6}\Loss_{\mathit{kl}},
\end{equation*}
where $\lambda_{1,\ldots ,6}$ are hyperparameters representing the relevance of each loss term.\looseness=-1
\subsection{Hyperparameter Optimization}\label{sec:approach_hpo}
Since observational statistics of a robot depend greatly on the specific environment that it collects data from and its onboard sensors, the best hyperparameter configuration of a model usually varies across different datasets. Manual hyperparameter optimization (HPO) is thus cumbersome and requires expert knowledge. To overcome this bottleneck, we equip our model with automated HPO (AutoML) techniques~\cite{hutter2019automated}, which have been shown to determine well-performing hyperparameters automatically. They improve performance substantially over manual tuning not just in supervised learning \cite{Feurer2019} but also in Model-based Reinforcement Learning and Planning \cite{zhang2021importance} which is closer to our setting. More concretely, we tune our model for automatically setting the learning rate, $\alpha$ (L1 or L2 reconstruction loss), and the $\lambda_{1, \ldots , 6}$ hyperparameters (importance weights of losses). We tested the Ray Tune \cite{moritz2018ray} implementations of ASHA \cite{li2020system}, BOHB \cite{falkner2018bohb} and HEBO~\cite{cowen2020hebo} and selected ASHA as it fully utilized the available parallel resources. ASHA is an asynchronous version of HyperBand \cite{hyperband_jmlr_17}, a multi-fidelity HPO approach. As such, it terminates runs with poorly performing configurations after having run for smaller budgets while promoting well performing configurations to run for larger budgets.\looseness=-1

\label{sec:approach}
\section{Experimental Evaluation}\label{sec:experiments}
We evaluate T3VIP for RGB-D video prediction on both simulated and real-world datasets. The goals of these experiments are to investigate: (i) whether our unsupervised generative model can make long-range RGB-D predictions via reasoning about the 3D dynamics of the scene, (ii) how our model leverages AutoML techniques to find a hyperparameter configuration that uses available observational cues, and (iii) if our generative model can be used as a 3D world model to plan for action trajectories that move a robot to user-defined 3D goal points.\looseness=-1
\subsection{Evaluation Metrics}
Following prior work on stochastic video prediction evaluation~\cite{babaeizadeh2018stochastic}, we sample 100 latent values from the prior distribution per video and report the best sample results. More concretely, for the predicted RGB images, we report PSNR and SSIM and VGG cosine similarity~\cite{zhang2018unreasonable} scores. VGG similarity has been shown to coincide better with human perceptual judgments~\cite{zhang2018unreasonable}. We further measure the standard quality metrics of predicted depth maps~\cite{eigen2015predicting} and report two widely-accepted error metrics: root mean square error (RMSE) and absolute relative error (AbsRel). Finally, we quantitatively evaluate the success rate of a model-predictive control algorithm that employs our 3D-aware model to plan for reaching 3D goal points.\looseness=-1
\subsection{Datasets}
We evaluate our model extensively with one real-world and two synthetic datasets. Please note that we use all datasets with the spatial resolution of $64 \times 64$. Datasets are as following:\looseness=-1

\noindent\textbf{DexHand}: Inspired by OpenAI dexterous in-hand manipulation~\cite{andrychowicz2020learning}, we collected a synthetic RGB-D dataset of a Shadow Hand robot manipulating a cube towards arbitrary goal configurations. This dataset consists of about $10000$ videos, each video including $25$ RGB-D frames. DexHand is challenging as the robot has $24$ degrees of freedom, and there can be a significant amount of motion and occlusion between consecutive frames.\looseness=-1

\noindent\textbf{CALVIN}~\cite{mees2021calvin}: This synthetic dataset includes $24$ hours of unstructured play data collected via teleoperating a Franka Emika Panda robot arm to manipulate objects in four visually distinct 3D environments. CALVIN is specifically interesting for us as it allows us to train our predictive model in one environment and test its invariance to object appearance in another unseen environment. Moreover, we use the CALVIN environment to perform model-predictive control via the learned 3D world model and plan for action trajectories. Concretely, our train and validation sets are from the CALVIN environment \textit{Env C}, and our test set is from \textit{Env D}.\looseness=-1

\noindent\textbf{Omnipush}~\cite{bauza2019omnipush}: Although there has been a considerable effort in the robotics community to collect real-world robot interactions~\cite{finn2016unsupervised, dasari2019robonet}, a significant limitation is the lack of depth modality in these datasets. To the best of our knowledge, Omnipush is the only real-world dataset that provides RGB-D videos recorded via a static camera looking towards the workspace of a robot pushing differently shaped objects. Omnipush is a challenging dataset as it consists of noisy actions and observations and reflects the stochastic dynamics of the real world very well. We use the first dataset split, consisting of 70 objects without extra weight. We use 50 objects for the training set, 10 for validation, and 10 for test sets.\looseness=-1
\begin{figure}[t]
	\centering
	\includegraphics[width=0.75\columnwidth]{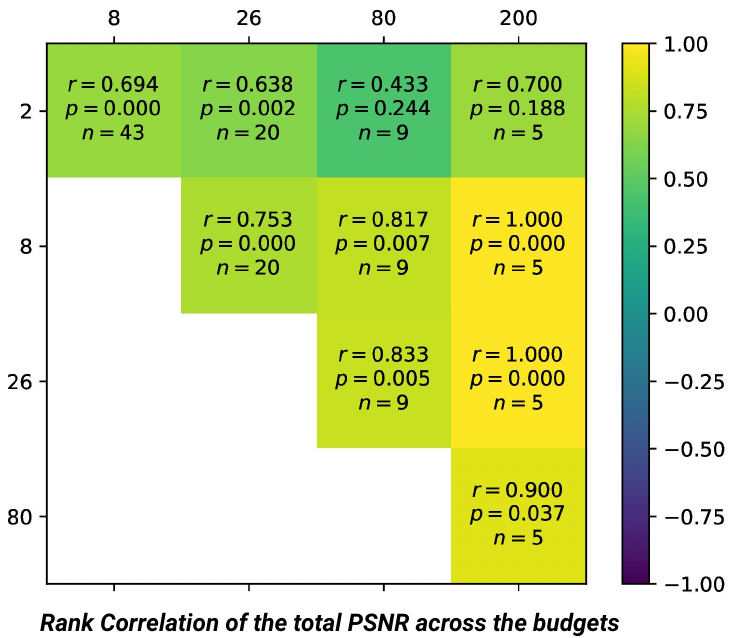}
	\caption{Spearman rank correlation across pairs of different budgets in epochs: The rank correlation for a pair of budgets can be read in the respective cell for a given row and column. While $r$ represents the correlation value and $p$ the p-value,  $n$ is the number of configurations that were common to the given pair of budgets. For instance, the rank correlation between the pair of budgets $2$ and $200$ can be read in the cell corresponding to the first row and fourth column as $0.700$ with a p-value of $0.188$ and $5$ common configurations that were promoted from the budget of $2$ epochs to the budget of $200$ epochs. As can be seen, the correlations across different pairs of budgets are fairly high and positive, implying that multi-fidelity HPO is efficient at finding well-performing configurations in our experiments.\looseness=-1}
	\label{fig:hpo_corr}
	\vspace*{-8mm}
\end{figure}
\begin{figure*}[t!]
    \centering
    \includegraphics[scale=0.64]{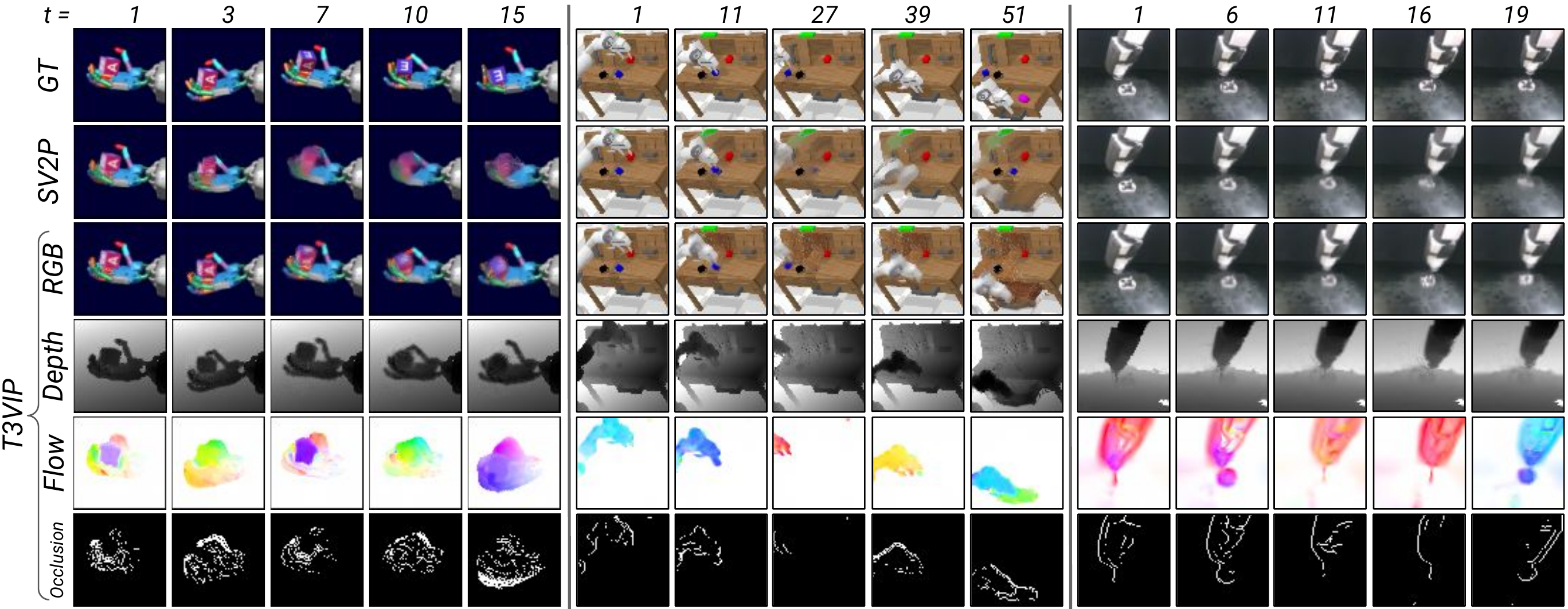}
    \caption{Qualitative results of T3VIP compared with SV2P~\cite{babaeizadeh2018stochastic} on DexHand, CALVIN, and Omnipush datasets, respectively. T3VIP predicts long-range RGB-D frames and outputs sharp scene and optical flows and sparse binary occlusion masks as emergent properties. T3VIP deals better with occlusions and generates sharper RGB images than the baseline. Please use the color wheel \raisebox{-0.10cm}{\includegraphics[height=0.35cm]{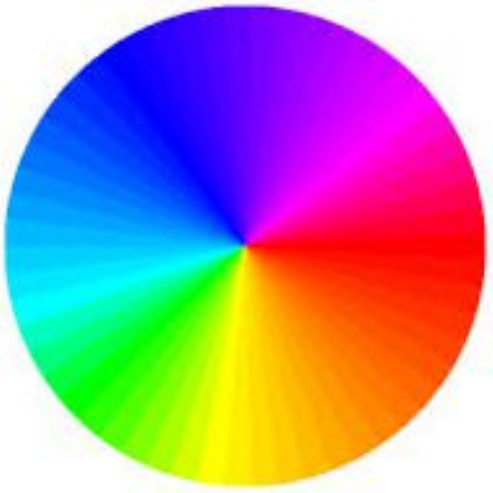}} to validate how our computed occlusion mask is compliant with our predicted flow field, \eg when the robot goes to the right (reddish color), the occlusion mask indicates that the robot's left side has been occluded before and now is visible and should get inpainted. A flow field with several distinct colors shows that our model decomposes the scene into objects and reasons about their 3D motion separately.\looseness=-1}
    \label{fig:qual}
    \vspace*{-8mm}
\end{figure*}
\vspace*{-6mm}
\subsection{Hyperparameter Optimization}\label{sec:expt_hpo}
As mentioned in \secref{sec:approach_hpo},  we employ ASHA~\cite{li2020system} separately on each dataset to automatically find hyperparameter configurations that lead to high-quality RGB-D predictions on that dataset. Thus, we consider the sum of PSNR scores for both the predicted RGB images and depth maps as the optimization metric of ASHA. We observed that the hyperparameter configurations found on smaller budgets still perform well after training for the full budget, and this allowed us to perform HPO much cheaper than using the full budget for every training run. The analysis of the HPO for the experiment on the Omnipush dataset in \figref{fig:hpo_corr} shows that the Spearman rank correlation across the different pairs of budgets is quite high (often 0.7 and above).
This intuitively explains the performance transfer from the max budget given to the HPO to the full budget of the final evaluation. Across all three datasets, we observed that ASHA finds not only one but several configurations that result in learning the 3D scene dynamics and consequently accurate RGB-D video predictions.\looseness=-1
\subsection{Comparisons and Ablations}
In our experiments, we compare our stochastic RGB-D video prediction model to the following well-established RGB video prediction baselines:\looseness=-1

\noindent\textbf{CDNA}: Deterministic video prediction model proposed in~\cite{finn2016unsupervised} that predicts the motion of pixels and transforms them from previous images to construct future images. CDNA has a similar architecture as our forward dynamics model, with the difference that it predicts motion via 2D convolution kernels and our model explicitly predicts 3D rigid body transformations.\looseness=-1

\noindent\textbf{SNA}: Inspired by~\cite{ebert2017self}, we empower the CDNA baseline to address occlusions by adding a residual at each time step via a skip connection from the first image in the video sequence and an additional predicted mask.\looseness=-1

\noindent\textbf{SV2P}: Stochastic video prediction model proposed in~\cite{babaeizadeh2018stochastic} which employs the deterministic CDNA model to perform next frame generation, but this time conditioned on stochastic latent variables sampled from a prior distribution. In this comparison, we adapted SV2P to use SNA instead of CDNA to enable it to reason about the occlusions.\looseness=-1

We follow the training recipe of~\cite{babaeizadeh2018stochastic} and train all models for $200K$ steps. We condition all the models on the first two frames and train them to predict the next ten frames. Quantitative results of each model in predicting future frames conditioned on the robot's actions are reported in ~\tabref{tab:quan-eval}. Our T3VIP successfully makes long-range RGB-D predictions across all three datasets via
reasoning about the 3D motion, while the baselines cannot reason about the scene's geometry. Furthermore, our model significantly outperforms baselines on the DexHand dataset (where much self-occlusion is present) in the RGB metrics while achieving on-par performance on CALVIN and Omnipush datasets. Although SNA shows improved performance compared to CDNA in handling occlusions, our model analytically computes a binary occlusion mask based on the predicted 3D motion and has superior performance in handling the missing regions of the scene. We validated our superior occlusion handling also on Omnipush and CALVIN by always skipping two frames in the datasets between the frames used for training the models. This skipping causes more motion and consequently more occlusion between consecutive frames, and~\tabref{tab:quan-eval} shows that our model outperforms the baselines in such scenarios.\looseness=-1

Our qualitative evaluation (displayed in~\figref{fig:qual}) shows that T3VIP produces sharp flow fields and sparse occlusion masks. This indicates that our model learns to reason about the 3D dynamics of objects in the scene and generates the next frame primarily using the predicted 3D motion and only inpainting those sparse occluded regions. Besides predicting long-range future depth maps, our model also produces sharper RGB images than baselines. The main reason for this is that beyond the reconstruction loss, our model also encourages sharp scene flow and optical flow fields.\looseness=-1
\bigbreak
\begin{table}[b!]
\vspace*{-10mm}
\footnotesize
\centering
\setlength\tabcolsep{1.3pt}
\renewcommand{\arraystretch}{0.95}
 \begin{tabular}{c|c|c|ccc|cc}
 \toprule
 \multirow{2}{*}{\centering \textbf{Dataset}} & \textbf{Test} & \multirow{2}{*}{\centering \textbf{Model}} & \multicolumn{3}{c|}{\cellcolor{red!25}\textbf{RGB}} & \multicolumn{2}{c}{\cellcolor{blue!25}\textbf{Depth}}\\
 & \textbf{Steps} & & \cellcolor{red!25}\textbf{PSNR}$\uparrow$ & \cellcolor{red!25}\textbf{SSIM}$\uparrow$ & \cellcolor{red!25}\textbf{VGG}$\uparrow$ & \cellcolor{blue!25}\textbf{\space RMSE}$\downarrow$ & \cellcolor{blue!25}\textbf{\space AbsRel}$\downarrow$\\
 \midrule
\parbox[t]{2mm}{\multirow{4}{*}{\rotatebox[origin=c]{90}{\parbox{1.0cm}{\centering DexHand}}}} & \multirow{4}{*}{\centering 25}  &
 CDNA~\cite{finn2016unsupervised} & $20.254$ & $0.803$ & $0.825$  & $\times$ & $\times$\\
 & & SNA~\cite{ebert2017self} & $20.643$ & $0.814$ & $0.836$  & $\times$ & $\times$\\
 & & SV2P~\cite{babaeizadeh2018stochastic} & $20.892$ & $0.810$ & $0.836$  & $\times$ & $\times$\\
 & & \ourmodel (Ours) & $\textbf{23.153}$ & $\textbf{0.889}$ & $\textbf{0.909}$  & $\textbf{0.024}$ & $\textbf{0.005}$\\
 \midrule
 \midrule
 \parbox[t]{2mm}{\multirow{4}{*}{\rotatebox[origin=c]{90}{\parbox{1.2cm}{\centering CALVIN $skip=0$}}}}&  \multirow{4}{*}{\centering 60}  &
 CDNA~\cite{finn2016unsupervised} & $22.374$ & $0.847$ & $0.875$  & $\times$ & $\times$\\
 & & SNA~\cite{ebert2017self} & $22.279$ & $0.842$ & $0.881$  & $\times$ & $\times$\\
 & & SV2P~\cite{babaeizadeh2018stochastic} & $22.363$ & $0.844$ & $\textbf{0.882}$  & $\times$ & $\times$\\
 & & \ourmodel (Ours) & $\textbf{22.692}$ & $\textbf{0.849}$ & $0.857$  & $\textbf{0.133}$ & $\textbf{0.006}$\\
 \midrule
 \midrule
 \parbox[t]{2mm}{\multirow{4}{*}{\rotatebox[origin=c]{90}{\parbox{1.2cm}{\centering CALVIN $skip=2$}}}}&  \multirow{4}{*}{\centering 60}  &
 CDNA~\cite{finn2016unsupervised} & $19.397$ & $0.740$ & $0.779$  & $\times$ & $\times$\\
 & & SNA~\cite{ebert2017self} & $19.403$ & $0.739$ & $0.787$  & $\times$ & $\times$\\
 & & SV2P~\cite{babaeizadeh2018stochastic} & $19.408$ & $0.741$ & $0.791$ & $\times$ & $\times$\\
 & & \ourmodel (Ours) & $\textbf{21.208}$ & $\textbf{0.779}$ & $\textbf{0.801}$  & $\textbf{0.125}$ & $\textbf{0.007}$\\
 \midrule
 \midrule
 \parbox[t]{2mm}{\multirow{4}{*}{\rotatebox[origin=c]{90}{\parbox{1.2cm}{\centering Omnipush $skip=0$}}}} & \multirow{4}{*}{\centering 60}  &
 CDNA~\cite{finn2016unsupervised} & $\textbf{25.239}$ & $0.877$ & $0.848$  & $\times$ & $\times$\\
 & & SNA~\cite{ebert2017self} & $23.855$ & $0.880$ & $0.871$  & $\times$ & $\times$\\
 & & SV2P~\cite{babaeizadeh2018stochastic} & $23.853$ & $\textbf{0.884}$ & $0.874$  & $\times$ & $\times$\\
 & & \ourmodel (Ours) & $24.261$ & $0.879$ & $\textbf{0.893}$  & $\textbf{0.053}$ & $\textbf{0.023}$\\
 \midrule
 \midrule
 \parbox[t]{2mm}{\multirow{4}{*}{\rotatebox[origin=c]{90}{\parbox{1.2cm}{\centering Omnipush $skip=2$}}}} & \multirow{4}{*}{\centering 20}  &
 CDNA~\cite{finn2016unsupervised} & $25.304$ & $0.890$ & $0.865$  & $\times$ & $\times$\\
 & & SNA~\cite{ebert2017self} & $25.452$ & $0.903$ & $0.891$  & $\times$ & $\times$\\
 & & SV2P~\cite{babaeizadeh2018stochastic} & $25.730$ & $\textbf{0.907}$ & $0.891$  & $\times$ & $\times$\\
 & & \ourmodel (Ours) & $\textbf{26.053}$ & $0.899$ & $\textbf{0.911}$  & $\textbf{0.050}$ & $\textbf{0.020}$\\
 \bottomrule
 \end{tabular}
\caption{Comparison between T3VIP and 2D baselines on average scores over the sequence of predicted videos. Please note that although all models are trained to predict the next ten time steps, we test them for longer horizons.}
\label{tab:quan-eval}
\end{table}
\begin{figure*}[t]
    \centering
    \subfloat[Evaluation of a 3D servoing experiment conducted via a 2D (SV2P) and a 3D world model (T3VIP)]{
        \includegraphics[scale=0.67]{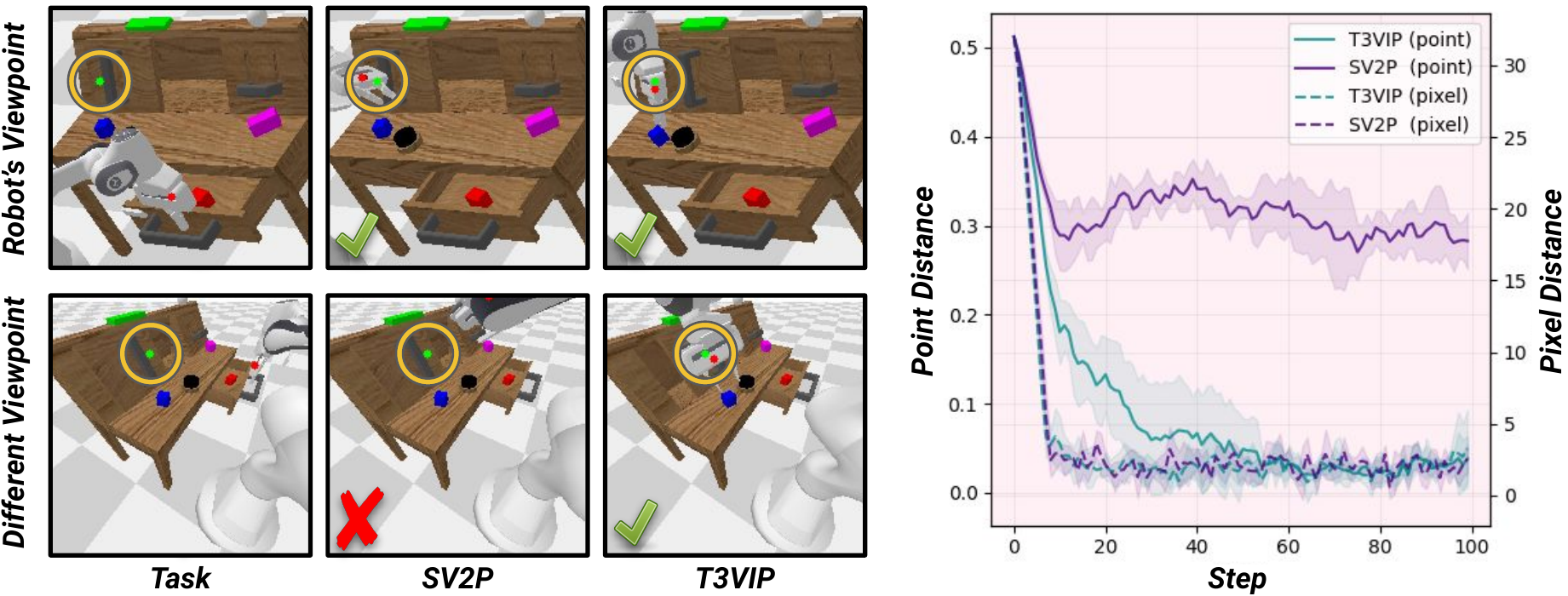}\label{fig:plan_scenario}}
    \subfloat[Overall success rate on 2D and 3D domains]{
        \includegraphics[scale=0.67]{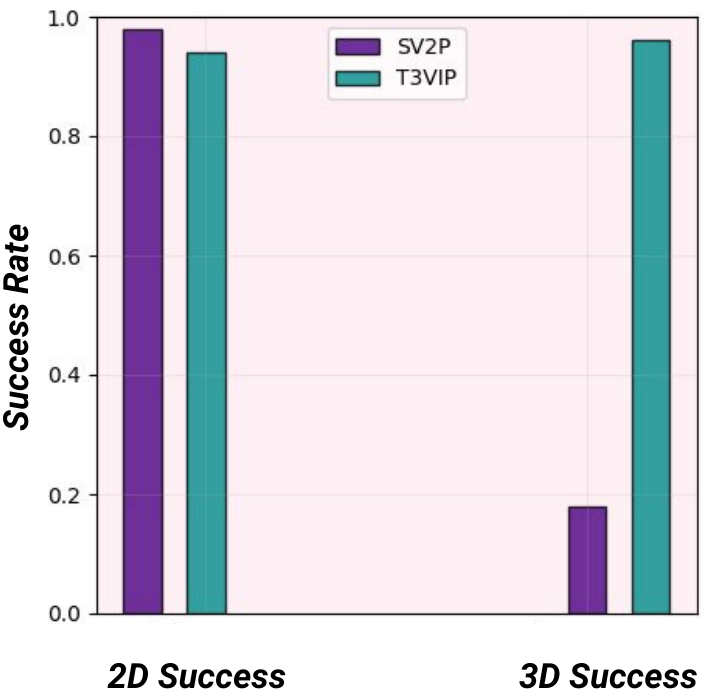}\label{fig:overall_success}}
  	\caption{(a) The first row visualizes the robot's viewpoint, whereas the second row visualizes an alternative view (the robot does not have access to this view). We observe that while SV2P reaches the pixel coordinates of the target, it misses the target 3D point considerably. However, T3VIP which is 3D-aware successfully reaches the target and solves the task in both 2D and 3D domains. (b) T3VIP significantly outperforms SV2P in 3D servoing experiments.}
	\label{fig:planning}
    \vspace{-0.8cm}
\end{figure*}

To analyze the influence of our different loss functions on the learned 3D dynamics and quality of RGB-D video prediction, we conducted an ablation experiment on the Omnipush dataset (see~\tabref{tab:ablation}). Our results indicate that a model (M1) that only aims to reduce the reconstruction loss is not effective in learning the 3D dynamics of the scene. Furthermore, a model that additionally regularizes the scene flow to be smooth (M2) was unsuccessful in reasoning about the scene dynamics. Although the third variant of our model (M3) that leverages point cloud alignment loss to enforce geometric consistency learns the dynamics of the scene, it leads to non-smooth flow fields. This is expected because the nearest-neighbor data association is prone to spurious matches, especially with noisy depth measurements. We found that variants that utilize 3D point cloud alignment and regularize for scene flow (M4) and both optical and scene flow fields (M5) lead to better reasoning about the dynamics and sharper predicted frames. Finally, our final model (M6) that also reasons about the stochasticity and minimizes the KL divergence between the approximated posterior and the assumed prior leads to the best RGB-D prediction performance.\looseness=-1

\begin{figure}[h]
	\vspace*{-3mm}
	\centering
	\includegraphics[width=0.65\columnwidth]{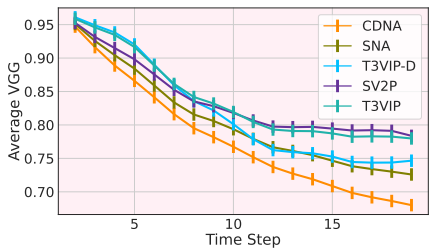}
	\caption{Comparison between T3VIP and baselines on action-free prediction of frames on Omnipush dataset. T3VIP-D is a deterministic variant of our model.\looseness=-1}
	\label{fig:act_free}
	\vspace*{-5mm}
\end{figure}

We also evaluated our model performance on Omnipush dataset without conditioning on actions, see~\figref{fig:act_free}. We observed that although the performance degrades compared to the action-conditioned setting, our model can still reasonably predict the future RGB-D frames. More specifically, we demonstrate that our full model, which captures stochasticity, outperforms a deterministic version of our model (T3VIP-D).\looseness=-1
\begin{table}[t]
\vspace*{2mm}
\centering
\footnotesize
\setlength\tabcolsep{0.05pt}
 \begin{tabular}{ccccccc|ccc|cc}
 \toprule
    \centering \textbf{Model} & \textbf{$\Loss_{rec}^{I}$} & \textbf{$\Loss_{rec}^{D}$}  & \textbf{$\Loss_{knn}$} & \textbf{$\Loss_{fs}^{s}$} & \textbf{$\Loss_{fs}^{o}$} & \textbf{$\Loss_{kl}$} & \cellcolor{red!25}\textbf{PSNR}$\uparrow$ & \cellcolor{red!25}\textbf{SSIM}$\uparrow$ & \cellcolor{red!25}\textbf{VGG}$\uparrow$ & \cellcolor{blue!25}\textbf{RMSE}$\downarrow$ &
    \cellcolor{blue!25}\textbf{AbsRel}$\downarrow$\\
 \midrule
 M1 & \checkmark & \checkmark & - & - & - & - & $\times$ & $\times$ & $\times$ & $\times$ & $\times$\\ 
 M2 & \checkmark & \checkmark & - & \checkmark & - & - & $\times$ & $\times$ & $\times$ & $\times$ & $\times$\\ 
 M3 & \checkmark & \checkmark & \checkmark & - & - & - & $24.92$ & $0.88$ & $0.86$ & $0.06$ & $0.02$\\ 
 M4 & \checkmark & \checkmark & \checkmark & \checkmark & - & - & $25.29$ & $0.88$ & $0.89$ & $0.05$ & $0.02$\\ 
 M5 & \checkmark & \checkmark & \checkmark & \checkmark & \checkmark & - & $25.53$ & $0.89$ & $0.90$ & $0.05$ & $0.02$\\
 M6 & \checkmark & \checkmark & \checkmark & \checkmark & \checkmark & \checkmark & $\textbf{26.05}$ & $\textbf{0.90}$ & $\textbf{0.91}$  & $\textbf{0.05}$ & $\textbf{0.02}$\\
 \bottomrule
 \end{tabular}
\caption{Ablation study: Our full model leverages all observational cues to reason about the 3D dynamics of the scene and predict future RGB-D frames.\looseness=-1}
\label{tab:ablation}
\vspace*{-2mm}
\end{table}
\subsection{Model-Predictive Control}
To evaluate the effectiveness of our 3D-aware world model and compare it with a 2D world model (SV2P), we conduct a 3D servoing experiment in the CALVIN environment. The goal of this experiment is for the robot arm to reach the user-defined goal point as closely as possible. To this end, we employ iCEM~\cite{pinneri2020sample}, which utilizes the learned world model to search for the best action trajectory that leads to the desired goal point. Please note that although the goal point is in 3D, a 2D world model is unaware of the target's depth measurement and can only comprehend the corresponding target pixel. Our model uses its predicted transformed point cloud to track the progress towards the target point, whereas SV2P uses its predicted 2D convolutional kernels to track the progress towards the target pixel. We use a discounted sum of trajectory distances to the target (point distance for T3VIP and pixel distance for SV2P) as the cost function. To conduct an extensive evaluation of the performances of T3VIP versus SV2P, we created a series of 10 experiment setups, where each setup consists of a random configuration of the scene and robot arm, a random goal point in 3D, and its corresponding pixel coordinates. We executed each experiment setup five times for each model with a fixed budget of 100 steps. As the distance of the robot end-effector to the goal point varies for each experiment setup, we consider an experiment successful if the end-effector reaches within 10 percent of the distance from the goal and meets this condition at least five times within an experiment. We observe that although SV2P can successfully reach the pixel coordinates of the goal point, it often misses reaching the user-defined 3D target (see~\figref{fig:plan_scenario}), as it suffers from the inherent ambiguity of 2D vision in capturing depth. In contrast, our model comprehends the goal in 3D, effectively predicts the 3D dynamics of the robot's end-effector, and successfully reaches the target point. \figref{fig:overall_success} shows the overall success of both models in reaching the target in 2D and 3D domains across all the experiments. Videos of these experiments are available at \url{http://t3vip.cs.uni-freiburg.de}.\looseness=-1

\newpage
\section{Conclusions}\label{sec:conclusion}
In this paper, we proposed transformation-based 3D video prediction (T3VIP) as a multi-step RGB-D video prediction approach that explicitly models the 3D dynamics and the rigid transformations of a scene. The unsupervised formulation of our approach leverages visual and geometric cues in the environment to learn real-world stochastic dynamics without human supervision. We also employ automated machine learning techniques to aid our model in finding hyperparameter configurations that exploit observational signals and achieve high accuracy. In extensive experiments we demonstrate that T3VIP learns an intuitive 3D world model, which outputs interpretable scene and optical flow fields and effectively enables an agent to reach 3D targets.\looseness=-1

Regarding future research, we believe that, given the ability to learn long-range RGB-D predictions of the future from unlabeled experience and the advanced capabilities of a 3D model in contrast to a 2D model, employing 3D models is a promising direction to autonomously learn real-world robot skills.\looseness=-1


\footnotesize
\bibliographystyle{IEEEtran}
\bibliography{references.bib}

\end{document}